\documentclass{article}

\PassOptionsToPackage{numbers, compress}{natbib}

\usepackage[preprint]{neurips_2026}

\usepackage[utf8]{inputenc} 
\usepackage[T1]{fontenc}    
\usepackage{graphicx}       
\usepackage{placeins}
\usepackage{booktabs}       
\usepackage{amsmath,amssymb}
\usepackage{enumitem}
\usepackage{amsfonts}       
\usepackage{multirow}
\usepackage{colortbl}
\usepackage{pifont}
\usepackage{subcaption}
\usepackage{nicefrac}       
\usepackage{microtype}      
\usepackage{xcolor}         
\usepackage{hyperref}       
\hypersetup{hidelinks}
\usepackage{url}            
\usepackage[capitalize]{cleveref}

\newcommand{\ours}{DexHLM}

\providecommand{\Description}[1]{}
\newcommand{\cmark}{\ding{51}}

\title{DexHLM: An Egocentric Dexterous Human Loco-Manipulation Dataset for Humanoids with Tactile Prediction}

\author{%
Zhenyu Wei$^{1,2,}$\thanks{\small Equal contribution. $\dagger$Corresponding authors.}\hspace{1.9em}%
Haoyang Luo$^{1,3,*}$\hspace{1.5em}%
Chixuan Zhang$^{1,3}$\hspace{1.5em}%
Guo Chen$^{1,3}$\hspace{1.5em}%
Suting Ni$^1$ \\
\textbf{Siyuan Huang$^2$\hspace{1.5em}%
Baoxiong Jia$^{2,\dagger}$\hspace{1.5em}%
Jingya Wang$^{1,3,\dagger}$} \\[12pt]
$^1$ShanghaiTech University\hspace{1.5em}%
$^2$State Key Laboratory of General Artificial Intelligence, BIGAI \\
$^3$InstAdapt \\[6pt]
\texttt{\{weizhy2024, luohy12024, zhangchx12024\}@shanghaitech.edu.cn} \\
\texttt{\{chenguo2024, nist2024, wangjingya\}@shanghaitech.edu.cn} \\
\texttt{huangsiyuan@ucla.edu\quad baoxiongjia@g.ucla.edu}
}

\begin{document}

\maketitle

\begin{center}
    \captionsetup{type=figure}
    \centering
    \vspace{-15pt}
    \includegraphics[width=\linewidth]{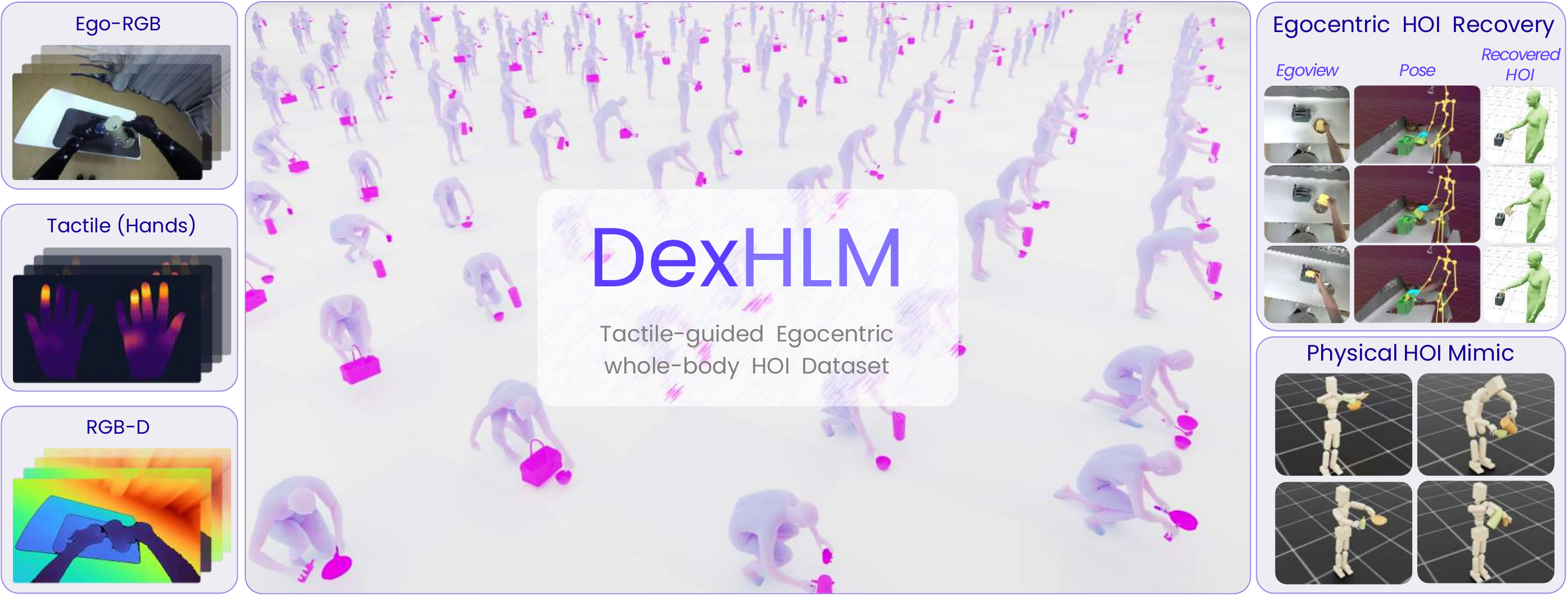}
    \caption{\textbf{Overview of \ours{}.} \ours{} is a multimodal egocentric HOI benchmark for humanoid whole-body dexterous manipulation, featuring synchronized tactile sensing, egocentric RGB-D, body motion, object trajectories, and hand motion across diverse spatial layouts and long-horizon activities.}
    \Description{Overview teaser of the DexHLM dataset, showing synchronized whole-body motion, object motion, egocentric RGB-D, tactile sensing, tactile-guided refinement, and egocentric HOI recovery.}
  \label{fig:teaser}
\end{center}

\begin{abstract}
  Humanoid whole-body loco-manipulation requires scalable and physically reliable demonstrations, yet existing human-object interaction (HOI) datasets lack direct contact supervision and egocentric observability, often resulting in physically inconsistent interactions. We present \ours{}, a multimodal egocentric HOI benchmark for humanoid whole-body dexterous manipulation, featuring synchronized tactile sensing, egocentric RGB-D, body motion, object trajectories, and hand motion across diverse spatial layouts and long-horizon activities. To improve physical consistency, we introduce a tactile-guided refinement pipeline that leverages measured pressure to correct hand-object interactions while preserving global motion fidelity. We further establish an Egocentric HOI Recovery method for tactile-guided reconstruction from first-person observations, along with an egocentric tactile prediction model that supplies contact evidence at recovery time. Together, our work provides a scalable and physically grounded foundation for learning humanoid whole-body dexterous interaction, pointing toward low-cost egocentric data collection for embodied intelligence.
  Project page is available at: \url{https://dexhlm.github.io/}.
\end{abstract}

\section{Introduction}
\label{sec:intro}

Humanoid whole-body loco-manipulation requires the tight integration of locomotion, postural balance, and fine-grained dexterous interaction. Learning such coordinated behaviors typically demands large-scale, high-quality demonstrations~\cite{xu2025intermimic}, yet collecting them on humanoid robots is costly and difficult to scale due to hardware, safety, and teleoperation constraints~\cite{Fu2024MobileAloha}. HOI data offers a promising alternative, since everyday human activities naturally contain rich whole-body interaction relevant to humanoid skill acquisition. To serve as physically grounded supervision for robot learning, however, such data must capture contact- and force-aware hand-object interactions, span diverse object layouts and spatial configurations, include long-horizon activities, and preserve egocentric observations aligned with robot perception. Due to these rigorous requirements on data quality, this type of large-scale, interaction-centric HOI data remains largely underexplored. 

Existing HOI datasets only partially satisfy these requirements. Hand-object datasets~\cite{contactpose_2020,dexycb,fan2023arctic} provide fine-grained manipulation data, primarily focusing on hand-object interaction. Whole-body HOI datasets~\cite{taheri2020grab,bhatnagar2022behave,huang2022intercap,araujo2023circle,li2023object,jiang2022chairs,zhang2024hoi,kim2025parahome,lu2025humoto} extend to body-scale interactions but typically infer contact from geometric proximity rather than direct sensing~\cite{zhou2022toch,xu2025interact}, leading to problems like hand-object penetration, floating fingers, and physically implausible grasps. Beyond pose capturing, most existing datasets cluster around fixed table-top setups or task-specific scenes, rarely revisiting the same interaction across heights, postures, and viewpoints that everyday activity actually spans. 

Based on these observations, we introduce \ours{}, a multimodal HOI dataset that jointly addresses the aforementioned requirements by \emph{(i)} capturing tactile information for contact- and force-aware interaction; \emph{(ii)} systematically varying interaction height, posture, and viewpoint for diverse spatial configurations; \emph{(iii)} including long-horizon activities with transitions across atomic interactions; and \emph{(iv)} providing egocentric observations and camera poses for future robot learning. Synchronized capture alone, however, does not guarantee physical reliability: body and object trajectories are stable, but finger-level contact remains fragile, and geometric proximity cannot distinguish fingers that truly carry contact from those merely close to the surface. We therefore introduce a tactile-guided refinement pipeline anchored by measured pressure, and further train RL policies on the refined demonstrations as a humanoid usability check.

To reduce the data acquisition cost for humanoid whole-body dexterous manipulation, we explore a low-cost egocentric data collection pipeline designed for human-to-humanoid transfer. The core challenge lies in obtaining high-quality 3D human-object interaction (HOI) data, particularly high-precision hand motion, from portable egocentric recordings, a setting where traditional motion capture and tactile gloves are unavailable. To address this, we build upon our \ours{} benchmark and propose an Egocentric HOI Recovery method with Tactile Refinement. Specifically, we first recover hand articulation, object pose, and body motion from egocentric observations using existing methods, then refine the physically implausible parts (e.g., hand-object penetration) via a tactile-guided optimization. Since tactile sensing is not accessible during deployment, we further introduce a VLM-based tactile predictor that estimates contact pressure from egocentric video clips. Experiments on \ours{} demonstrate that our approach significantly improves the physical fidelity of reconstructed interactions, validating the effectiveness of leveraging VLM-guided tactile prediction for low-cost, high-precision HOI data acquisition.
In summary, our contributions are:
\begin{itemize}[leftmargin=4pt, rightmargin=4pt]
  \item We present \ours{}, a multimodal egocentric HOI benchmark for humanoid whole-body dexterous manipulation. It provides synchronized whole-body motion, object trajectories,  hand motion, real tactile sensing, and egocentric RGB-D across diverse spatial layouts and long-horizon activities.

  \item We introduce a tactile-guided refinement pipeline that leverages measured pressure to correct hand-object interactions (e.g., penetration and unsupported contacts) while preserving global motion fidelity, significantly improving the physical reliability of dexterous whole-body motion captures.

  \item We establish an Egocentric HOI Recovery method on \ours{} for tactile-guided HOI reconstruction from first-person observations, together with a VLM-based tactile prediction model that supplies contact evidence at recovery time, enabling downstream physical refinement without requiring measured tactile during deployment.
\end{itemize}

\begin{figure}[t]
  \centering
  \includegraphics[width=\linewidth]{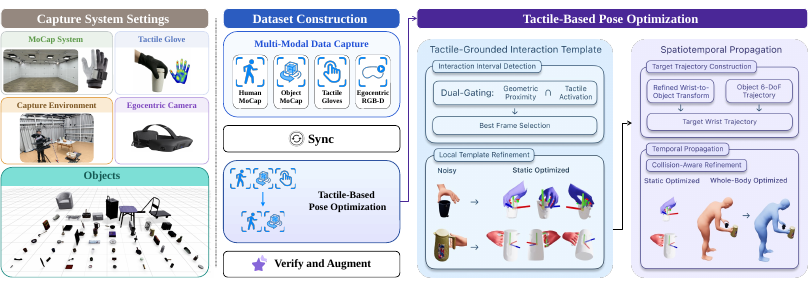}
  \caption{\textbf{Dataset construction and tactile-guided post-optimization.} \ours{} aligns whole-body motion, object motion, egocentric RGB-D, real tactile sensing, and controlled spatial layouts; the refinement uses measured tactile to convert raw capture into physically grounded demonstrations.}
  \Description{Overview figure of the DexHLM dataset, showing synchronized capture hardware, multimodal data streams, controlled spatial layouts, and the tactile-guided refinement pipeline.}
  \vspace{-15pt}
  \label{fig:overview}
\end{figure}

\section{Related Work}
\label{sec:related}

\noindent\textbf{Human-Object Interaction Datasets.}

Hand-object datasets such as ContactPose~\cite{contactpose_2020}, DexYCB~\cite{dexycb}, H2O~\cite{kwon2021h2o}, and ARCTIC~\cite{fan2023arctic} primarily focus on fine-grained hand articulation under close-range or first-person views. Whole-body HOI datasets such as BEHAVE~\cite{bhatnagar2022behave}, InterCap~\cite{huang2022intercap}, OMOMO~\cite{li2023object}, and HOI-M$^3$~\cite{zhang2024hoi} capture body-scale interactions, yet typically retain coarse hand articulation and infer contact from geometric proximity rather than direct measurement~\cite{zhou2022toch,xu2025interact}. Datasets such as GRAB~\cite{taheri2020grab}, HUMOTO~\cite{lu2025humoto}, and ParaHome~\cite{kim2025parahome} approach whole-body dexterity, but remain limited by missing tactile evidence, missing synchronized egocentric observation, or insufficient systematic spatial variation. To bridge this gap, to our knowledge, \ours{} is the first whole-body HOI corpus to unify measured tactile contact, egocentric RGB-D, systematic spatial layouts, and long-horizon activity at dexterous whole-body resolution.

\noindent\textbf{Interaction Optimization and Refinement.}

Hand-level methods such as ContactOpt~\cite{grady2021contactopt}, reconstruction pipelines~\cite{cao2021handobject}, GraspTTA~\cite{jiang2021hand}, and TOCH~\cite{zhou2022toch} achieve fine-grained hand-object refinement; whole-body pipelines such as CONTHO~\cite{Nam_2024_CVPR}, InterDiff~\cite{xu2023interdiff}, and HOI-FHLI~\cite{hoifhli} extend contact reasoning to full-body motion yet rely on inferred geometric proximity or learned priors rather than measured contact; and dataset-level methods such as InterAct~\cite{xu2025interact} unify geometric correction across corpora yet lack direct tactile supervision. To bridge this gap, our approach grounds correction in measured tactile pressure, modulating contact attraction while softly preserving captured motion.

\noindent\textbf{Egocentric Interaction Recovery.}

Egocentric datasets~\cite{kwon2021h2o,Ego4D2022CVPR} provide rich first-person observability yet lack synchronized tactile measurements. H+O~\cite{tekin2019ho} estimates hand-object poses. ContactHands~\cite{narasimhaswamy2020detecting} studies contact-state recognition, PressureVision~\cite{grady2022pressurevision} estimates hand pressure, and OpenTouch~\cite{OpenTouch} provides synchronized egocentric video, tactile signals, and hand pose. To bridge egocentric observation and physical contact, our approach integrates tactile estimation into whole-body HOI recovery, where predicted contact evidence guides finger refinement and object-trajectory correction rather than remaining a standalone perception baseline.

\section{The \ours{} Dataset}
\label{sec:dataset}

\subsection{Data Collection}
\label{sec:overview}

\ours{} synchronizes whole-body SMPL-X~\cite{2019SMPLX} motion, rigid-body 6-DoF object trajectories, camera poses, head-mounted egocentric RGB-D, and per-frame tactile pressure from a pair of custom gloves, each with 256 sensing points, for humanoid whole-body loco-manipulation. The corpus spans 20.1 hours over 52 objects and six everyday scenes, with both single-handed and bimanual activities covering prehensile, in-hand, and non-prehensile behaviors. It includes six participants, four male and two female, with varying heights and body proportions. The modalities are temporally aligned at 30\,Hz, with human and object motion expressed in a shared coordinate system.

Two protocol-level choices distinguish \ours{} from existing HOI corpora. First, same-level interactions sample a grid of five height levels (L1 floor to L5 neck), calibrated to each participant's body proportions, and three horizontal variations (V1--V3), so posture, reach, and viewpoint vary independently of task semantics. Second, capture proceeds through three temporal stages: same-level interactions (S1) populate grid cells, cross-level transitions (S2) chain locomotion with manipulation, and long-horizon scene-level activity (S3) combines interactions in freely arranged layouts. S3 also provides synchronized exocentric views as a qualitative reference, not as recovery inputs. \Cref{tab:DexHLM_comparison} compares \ours{} with representative HOI datasets along the modality and protocol axes most relevant to egocentric whole-body recovery; further dataset details are in \cref{app:dataset_details}.

\begin{table}[t]
\caption{Representative HOI datasets grouped by interaction scope. Columns summarize available sensing modalities and protocol properties, including tactile sensing, egocentric observation, spatial diversity, long-horizon activity, and dexterous-level hand interaction. Durations are approximate, reported or estimated from temporal sampling statistics, and exclude duplicate camera views; N/R denotes an unreported total duration.}
\label{tab:DexHLM_comparison}
\centering
\small
\setlength{\tabcolsep}{4pt}
\resizebox{\linewidth}{!}{%
\begin{tabular}{@{}l ccccc cc@{}}
\toprule
Dataset
& Tactile
& Egocentric
& \shortstack{Spatial Diversity}
& \shortstack{Long Horizon}
& \shortstack{Dexterous Hand}
& \shortstack{Duration (h)}
& \shortstack{\# of Objects} \\
\midrule
\multicolumn{8}{@{}l}{\emph{Hand-object interaction}} \\
ContactPose~\cite{contactpose_2020}
& -- & -- & -- & -- & \cmark & N/R & 25 \\
DexYCB~\cite{dexycb}
& -- & -- & -- & -- & \cmark & 0.7 & 20 \\
H2O~\cite{kwon2021h2o}
& -- & \cmark & -- & -- & \cmark & 1.1 & 8 \\
ARCTIC~\cite{fan2023arctic}
& -- & \cmark & -- & -- & \cmark & 2.2 & 11 \\
\midrule
\multicolumn{8}{@{}l}{\emph{Whole-body HOI with coarse-hand annotation}} \\
BEHAVE~\cite{bhatnagar2022behave}
& -- & -- & -- & -- & -- & 4.2 & 20 \\
InterCap~\cite{huang2022intercap}
& -- & -- & -- & -- & -- & 0.6 & 10 \\
OMOMO~\cite{li2023object}
& -- & -- & -- & -- & -- & 10.0 & 15 \\
HOI-M$^3$~\cite{zhang2024hoi}
& -- & -- & -- & \cmark & -- & 20.0 & 90 \\
\midrule
\multicolumn{8}{@{}l}{\emph{Whole-body HOI with dexterous hand articulation}} \\
GRAB~\cite{taheri2020grab}
& -- & -- & -- & -- & \cmark & 3.8 & 51 \\
ParaHome~\cite{kim2025parahome}
& -- & -- & -- & \cmark & \cmark & 8.1 & 22 \\
HUMOTO~\cite{lu2025humoto}
& -- & -- & -- & \cmark & \cmark & 2.2 & 63 \\
\midrule
\rowcolor{gray!20}
\textbf{\ours{} (Ours)}
& \cmark & \cmark & \cmark & \cmark & \cmark & 20.1 & 52 \\
\bottomrule
\end{tabular}%
}
\vspace{-10pt}
\end{table}

\subsection{Measured-Tactile Human-Motion Refinement}
\label{sec:refine}

Body and object trajectories from MoCap are reliable, while local hand-object contact is the most fragile part of whole-body interaction data. Methods that infer contact from learned predictors or geometric proximity~\cite{grady2021contactopt,zhou2022toch,xu2023interdiff,xu2025interact} can confuse fingers that truly carry contact with those merely close to the surface, and over-correcting the latter disturbs the captured motion. We therefore anchor correction in measured tactile pressure, modulating contact attraction while softly constraining pose deviations. During dataset construction, the pipeline refines hand articulation and wrist pose against the reliable MoCap object trajectory, then propagates the correction to body posture. We describe these steps next, with additional details in \cref{app:tactirefine_full}; object-trajectory correction at egocentric recovery is described separately in \cref{sec:recovery_pipeline}.

\paragraph{Interval detection.}
A dual tactile-geometric gate flags frame $t$ when the maximum aggregated finger pressure exceeds $\tau_p$ and the hand-mesh-to-object distance falls below $\tau_d$. Within each smoothed interval, a representative template frame $t^*$ is chosen by maximizing a weighted score over tactile confidence, kinematic stability, and geometric proximity, defining the wrist-relative reference for the subsequent steps.

\paragraph{Pressure-modulated local refinement.}
At the template frame $t^*$, hand articulation $\boldsymbol{\theta}_{t^*}$ and wrist pose $\mathbf{T}_w$ are optimized under a per-finger objective whose weights derive from tactile evidence. We split the articulation into per-finger axis-angle blocks, $\boldsymbol{\theta}_{t^*}=\bigl(\boldsymbol{\theta}_{t^*}^{(1)},\dots,\boldsymbol{\theta}_{t^*}^{(F)}\bigr)$ with $\boldsymbol{\theta}_{t^*}^{(f)}\in\mathbb{R}^{3 J_f}$ stacking the axis-angle parameters of finger $f$'s $J_f$ joints. For each finger $f$, we form a tactile weight $w_f\in[0,1]$ by averaging measured fingertip pressure over a temporal window centered at $t^*$, normalizing to a global pressure scale, and clipping to $[0,1]$; the following simplified objective illustrates the balance between contact attraction and pose preservation:
\begin{equation}
  \min_{\boldsymbol{\theta}_{t^*},\,\mathbf{T}_w}\;\sum_{f=1}^{F}\!\Bigl[
     w_f\!\!\sum_{v\in\mathcal{V}_f}\!\bigl|\mathrm{SDF}_{\mathcal{O}}(\mathbf{v}_{t^*})\bigr|
     \;+\;(1-w_f)\,\bigl\lVert \boldsymbol{\theta}_{t^*}^{(f)} - \boldsymbol{\theta}_{t^*}^{(f),0}\bigr\rVert^2
  \Bigr],
  \label{eq:local_refine}
\end{equation}
where $\mathcal{V}_f$ is the index set of hand vertices on finger $f$, $\mathbf{v}_{t^*}$ denotes a hand vertex at frame $t^*$ expressed in the object's local frame, $\mathrm{SDF}_{\mathcal{O}}$ is the object signed distance field, and $\boldsymbol{\theta}_{t^*}^{(f),0}$ is the captured pose of finger $f$. The implementation uses Euclidean distances from sparse contact points to targets on the object surface, with attraction introduced in stages and additional regularizers (\cref{app:tactirefine_full}). Continuous weights balance attraction and soft pose preservation rather than freezing individual fingers.

\paragraph{Spatiotemporal propagation.}
The refined wrist-relative transform is composed with the object trajectory to obtain a target wrist trajectory $\hat{\mathbf{T}}_t$, blended with the captured wrist position $\mathbf{p}_t^{w,0}$ and finger pose $\boldsymbol{\theta}_t^{0}$ through smooth weights $\alpha_t^w,\alpha_t^f\in[0,1]$ that ramp inside each interval and decay at its boundaries:
\begin{equation}
  \mathbf{p}_t^w \;=\; (1-\alpha_t^w)\,\mathbf{p}_{t}^{w,0} + \alpha_t^w\,\hat{\mathbf{p}}_t,
  \qquad
  \boldsymbol{\theta}_t^{\text{blend}} \;=\; (1-\alpha_t^f)\,\boldsymbol{\theta}_t^{0} + \alpha_t^f\,\boldsymbol{\theta}^*,
  \label{eq:blend}
\end{equation}
with $\boldsymbol{\theta}^*$ the locally refined hand articulation. Body posture is propagated through the arm chain by bounded inverse kinematics with the root pose fixed, so contact frames track the refined wrist target while frames outside the interval stay close to the captured motion.

\paragraph{Collision-aware finger refinement.}
With the wrist trajectory fixed, finger pose on near-contact frames is refined by
\begin{equation}
  \min_{\boldsymbol{\theta}_t}
  \;\; w_{\text{pen}}\!\sum_{v} \kappa_v\,\mathrm{ReLU}\!\bigl(\mathrm{SDF}_{\mathcal{O}}(\mathbf{v}_t)+\delta\bigr)
  \;+\; w_{\text{close}}\,\bigl\lVert \boldsymbol{\theta}_t-\boldsymbol{\theta}_t^{\text{blend}}\bigr\rVert^2,
  \label{eq:collision}
\end{equation}
where $\mathbf{v}_t$ ranges over hand vertices in the object's local frame, $\mathrm{SDF}_{\mathcal{O}}$ is the object signed distance field positive inside the object, $\delta\!\geq\!0$ is a clearance margin, and $\kappa_v$ favors fingertip and palmar vertices. The closeness term anchors the result to the blended pose $\boldsymbol{\theta}_t^{\text{blend}}$, limiting drift from the tactile-guided correction.

\section{Egocentric HOI Recovery}
\label{sec:benchmark}

\subsection{Desiderata and Setup}
\label{sec:setup}

The Egocentric HOI Recovery Benchmark turns the question raised in \cref{sec:intro} into a concrete evaluation: which whole-body HOI states can be recovered from egocentric observations given GT camera poses, object meshes, and depth? We decompose whole-body HOI into three coupled states---hand articulation, object 6-DoF motion, and body pose---and let each egocentric method fill the column it can provide. Camera trajectory and object mesh are held at GT values, with wrist positions lifted using GT depth, so that metric variation reflects the component being evaluated rather than scene-reconstruction noise; tactile, by contrast, is treated as a derived signal predicted by \cref{sec:tactile_pred} and fed into the correction of \cref{sec:recovery_pipeline}, so the benchmark probes \emph{tactile-guided} egocentric recovery without requiring measured tactile at evaluation time. Hand and body errors use MoCap motion refined using measured tactile signals as reference; object errors use MoCap object trajectories.

The component baselines pair FoundationPose~\cite{wen2024foundationpose} with HaPTIC~\cite{haptic} for hand articulation or EgoEgo~\cite{li2023ego} for body pose. Our pipeline uses WiLoR~\cite{potamias2025wilor} for hands and EgoAllo~\cite{yi2025egoallo} for body pose (\cref{sec:recovery_pipeline}).

\begin{figure}[!t]
  \centering
  \includegraphics[width=\linewidth]{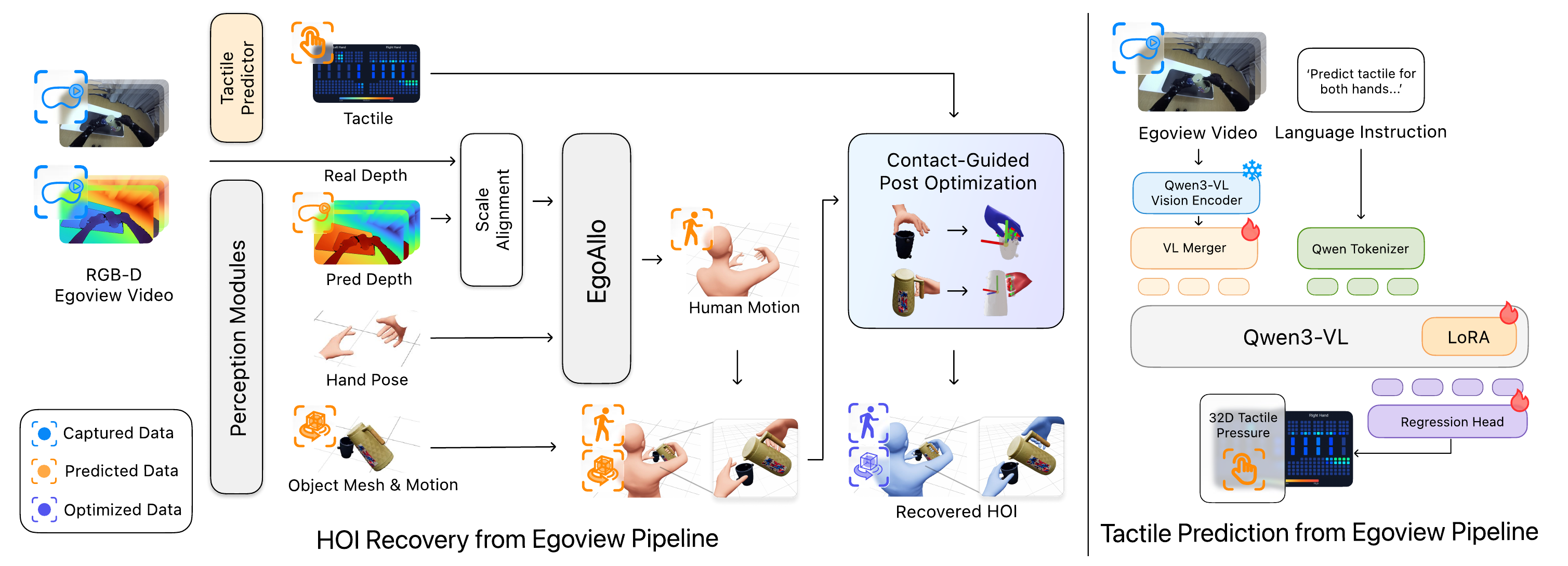}
  \caption{\textbf{Egocentric HOI Recovery Benchmark and tactile prediction.} The reported evaluation uses GT camera poses, GT object meshes, and wrists lifted using GT depth; predicted depth and mesh modules shown in the schematic are not evaluated here. Predicted tactile signals guide post-optimization at recovery time.}
  \Description{The egocentric HOI recovery pipeline and tactile predictor; evaluation uses GT camera poses, GT object meshes, and GT depth.}
  \label{fig:benchmark_egotacti}
\vspace{-15pt}
\end{figure}

\subsection{Tactile Prediction from Egocentric Video}
\label{sec:tactile_pred}

\paragraph{Problem formulation.}
The tactile predictor supplies the contact mask used by the object-trajectory correction of \cref{sec:recovery_pipeline}, so that recovery operates without a tactile glove at evaluation time. Given a short egocentric video clip and a fixed task-conditioning prompt, the model predicts a non-negative $32$-dimensional tactile representation for the target frame, associated with selected SMPL-X hand keypoints. Framewise predictions are thresholded into a binary contact mask for downstream object-trajectory correction.

\paragraph{Architecture.}
We adopt Qwen3-VL~\cite{Qwen3-VL} at 2B and 8B scales as the multimodal backbone. Because our curated egocentric--tactile corpus is orders of magnitude smaller than the backbone's web-scale pre-training data, we fine-tune through lightweight LoRA adapters on the language-layer attention projections instead of full-parameter updates, and attach a small two-layer MLP regression head inspired by OpenVLA~\cite{kim24openvla}. The head maps the backbone features to target-frame tactile values using synchronized glove measurements as supervision. We also train the visual-language merger and regression head while freezing the remaining backbone weights. Denoting a pre-trained weight matrix by $\mathbf{W}_0\in\mathbb{R}^{d\times k}$, the adapted forward pass is
\begin{equation}
\mathbf{h} \;=\; \mathbf{W}_0\,\mathbf{x} \;+\; \frac{\alpha}{r}\,\mathbf{B}\mathbf{A}\,\mathbf{x},\qquad \mathbf{A}\in\mathbb{R}^{r\times k},\;\;\mathbf{B}\in\mathbb{R}^{d\times r},
\label{eq:lora}
\end{equation}
with the adapter settings given in \cref{app:egotacti}.

\paragraph{Training objective.}
The predictor is trained on \ours{} using synchronized egocentric--tactile pairs. In the following general loss form, the temporal-smoothing weight is set to zero in our experiments:
\begin{equation}
\mathcal{L} \;=\; \mathcal{L}_{\text{MSE}} \;+\; \lambda_{\text{bce}}\,\mathcal{L}_{\text{BCE}} \;+\; \lambda_{\text{sm}}\,\mathcal{L}_{\text{smooth}}.
\label{eq:tactile_loss}
\end{equation}
The active terms combine weighted mean-squared error on tactile values, with non-zero targets up-weighted to counter contact sparsity, and binary cross-entropy on contact masks derived from a fixed pressure threshold.

\paragraph{Pipeline integration.}
Once trained, the predictor supplies contact evidence directly to the object-trajectory correction in \cref{sec:recovery_pipeline}, turning egocentric tactile estimation into an actionable signal for downstream HOI recovery (see \cref{sec:exp_tactile} for quantitative results and \cref{app:egotacti} for implementation details).

\subsection{Recovery with Predicted Tactile}
\label{sec:recovery_pipeline}

Hands and objects often appear in close proximity even when no interaction occurs, so geometric proximity is unreliable for correcting object trajectories. We therefore separate component-level recovery from tactile-guided post-processing: egocentric perception produces initial hand, body, and object estimates, and predicted contact intervals guide object-trajectory correction.

Hand articulation uses WiLoR~\cite{potamias2025wilor}, with each per-frame wrist lifted by the GT depth map. Body pose is recovered with EgoAllo~\cite{yi2025egoallo}, conditioned on GT camera poses and the depth-lifted wrist constraints. Object 6-DoF is recovered by FoundationPose~\cite{wen2024foundationpose}, complemented by four post-processing steps targeting the residual errors of the tracker: symmetric-orientation disambiguation, adaptive rotation-outlier rejection, sliding-window quaternion-mean smoothing, and multi-object secondary alignment.

\paragraph{Predicted-tactile object-trajectory correction.}
The recovered body and wrist trajectories are held fixed, while finger poses and the object trajectory are adjusted. The perception-time object estimate at the template frame directly anchors the correction; no iterative object-pose optimization is performed. For each detected contact interval $\mathcal{C}$, we record the wrist-relative object pose $T^{\mathrm{rel}}_{\mathrm{obj}\mid\mathcal{C}}\in SE(3)$ at the template frame and propagate it through the interval by composing it with the recovery-time wrist trajectory. Under approximately stable contact, this relation is expressed as $T_{\mathrm{obj}}(t)=T_{\mathrm{wrist}}(t)\,T^{\mathrm{rel}}_{\mathrm{obj}\mid\mathcal{C}}$ for $t\in\mathcal{C}$. The object trajectory can also change outside $\mathcal{C}$ through anchoring; finger poses remain subject to refinement. Predicted tactile signals identify contact intervals, rather than supplying the measured-pressure local weights of \cref{sec:refine}. Recovery thus operates without measured tactile at evaluation time (\cref{app:egotacti}).

\section{Experiments}
\label{sec:experiments}

\Cref{sec:exp_postopt} validates the tactile-guided refinement on raw \ours{} captures, \cref{sec:exp_tactile} evaluates the egocentric tactile predictor, \cref{sec:exp_recovery} reports the headline egocentric HOI recovery results, and \cref{sec:exp_humanoid} provides a preliminary humanoid usability check.

\subsection{Measured-Tactile Human-Motion Refinement}
\label{sec:exp_postopt}

\begin{table}[!htbp]
  \caption{Refinement ablation on raw \ours{} captures. FP-C, HPD, NTF-D, and MPJPE are not reported for the Raw row. Ours Full applies tactile-guided refinement combined with collision-aware finger refinement (\cref{sec:refine}). Metric definitions: \cref{app:metrics}.}
  \label{tab:postopt}
  \vspace{0.6em}
  \centering
  \footnotesize
  \setlength{\tabcolsep}{4pt}
  \begin{tabular}{@{}l ccccc@{}}
    \toprule
    Condition                & PR$\downarrow$(\%) & FP-C$\downarrow$(\%) & HPD$\downarrow$(rad) & NTF-D$\downarrow$(mm) & MPJPE$\downarrow$(mm) \\
    \midrule
    Raw MoCap                & 83.5         & ---           & ---            & ---           & ---           \\
    Geometry only            & 5.1          & 72.5          & 1.294          & 31.6          & 43.7          \\
    Tactile only             & 71.3         & 72.8          & 1.143          & 30.4          & 40.9          \\
    \textbf{Ours Full}       & \textbf{4.2} & \textbf{71.9} & \textbf{1.086} & \textbf{29.9} & \textbf{39.4} \\
    \bottomrule
  \end{tabular}
\end{table}

\begin{figure}[t]
  \centering
  \includegraphics[width=\linewidth]{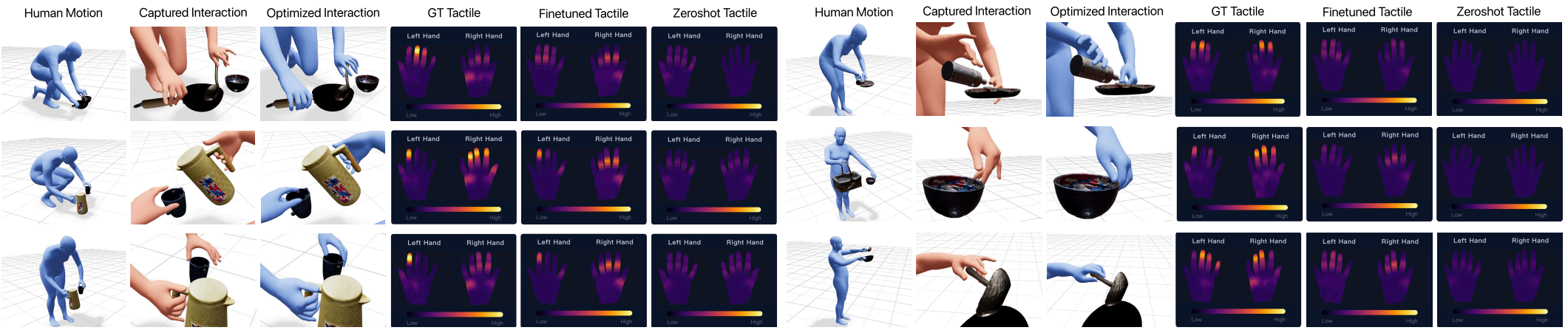}
  \caption{\textbf{Qualitative results of tactile-guided refinement and tactile prediction.} Tactile-guided refinement improves hand-object alignment while preserving natural articulation. The fine-tuned tactile predictor produces outputs closely matching measured tactile signals.}
  \Description{Qualitative comparison showing improved hand-object alignment after tactile-guided refinement and accurate tactile prediction from a fine-tuned tactile predictor.}
  \label{fig:qual_results}
\end{figure}

\Cref{tab:postopt} validates the refinement of \cref{sec:refine} on raw \ours{} captures along two axes: physical validity (PR, FP-C) and motion fidelity (HPD, NTF-D, MPJPE). Full refinement reduces the penetration-frame rate from $83.5\%$ to $4.2\%$, slightly below the geometry-only baseline ($5.1\%$), while also lowering FP-C ($71.9\%$ vs $72.5\%$) and keeping HPD, NTF-D, and MPJPE below the geometry-only baseline. The two ablated variants illustrate the balance between geometric correction and motion preservation: the geometry-only baseline reaches comparable penetration only at the cost of the largest motion deviation in the table (HPD $1.294$, NTF-D $31.6$\,mm, MPJPE $43.7$\,mm), and the Tactile-only variant without collision refinement leaves substantial penetration ($71.3\%$) despite using the same tactile-guided weighting. The full pipeline balances contact alignment with soft preservation of captured motion, consistent with \cref{sec:refine}; representative cases are shown in \cref{fig:qual_results}.

\subsection{Tactile Prediction from Egocentric Video}
\label{sec:exp_tactile}

\begin{table}[t]
  \caption{Egocentric tactile prediction. CF1: element-level contact F1; FC@$k$: frame-level contact F1. The zero-shot generation and invalid-output handling are described in \cref{app:egotacti}.}
  \label{tab:tactile_pred}
  \centering
  \vspace{0.35em}
  \footnotesize
  \setlength{\tabcolsep}{4pt}
  \begin{tabular}{@{}ll cccc cc c@{}}
    \toprule
                              & Model       & Pearson$\uparrow$ & CF1$\uparrow$  & FC@3$\uparrow$ & FC@5$\uparrow$ & MAE$\downarrow$ & MSE$\downarrow$ & R$^2$ $\uparrow$ \\
    \midrule
    \multirow{2}{*}{Zero-shot}
                              & 2B          & 0.000             & 0.000          & 0.000          & 0.000          & 3.006           & 33.22           & $-$0.42          \\
                              & 8B          & 0.320             & 0.435          & 0.431          & 0.468          & 2.776           & 25.34           & $-$0.18          \\
    \midrule
    \multirow{2}{*}{Ours}
                              & 2B          & 0.790             & 0.789          & 0.910          & 0.921          & 1.614           & 8.16            & 0.50             \\
                              & \textbf{8B} & \textbf{0.812}    & \textbf{0.816} & \textbf{0.925} & \textbf{0.922} & \textbf{1.449}  & \textbf{7.12}   & \textbf{0.60}    \\
    \bottomrule
  \end{tabular}
\end{table}

\Cref{tab:tactile_pred} measures whether the synchronized tactile of \ours{} makes egocentric tactile prediction a viable learning problem. The main evaluation allows seen participants and objects, including the same interactions at different heights. Under the zero-shot generation and output-handling protocol (\cref{app:egotacti}), the 2B model scores zero contact F1, and the 8B model yields only a weak signal (Pearson $0.32$, contact F1 $0.44$). Fine-tuning on \ours{} brings both scales into a usable regime (Pearson $0.79$/$0.81$, contact F1 $0.79$/$0.82$ for 2B/8B), and the 8B backbone delivers only marginal additional headroom over 2B. In the main evaluation, Ours-8B outperforms our supervised DINOv2~\cite{oquab2023dinov2} ViT-B/14 baseline, adapted from the design of TouchAnything~\cite{zhou2026touchanything} (baseline Pearson $0.674$, MAE $1.994$, MSE $12.45$). On jointly unseen participants and objects, \cref{tab:joint_holdout} compares both predictors without adaptation across all seven metrics, with Ours-8B reaching Pearson $0.795$. These results show that \ours{} supplies useful supervision for turning egocentric observations into contact evidence for the object-trajectory correction of \cref{sec:recovery_pipeline}. Baseline settings and additional results are in \cref{app:egotacti,app:additional_results}.

\subsection{Recovery with Predicted Tactile}
\label{sec:exp_recovery}

\begin{table*}[t]
\caption{Egocentric HOI recovery on \ours{} using GT camera poses, GT object meshes, and wrists lifted using GT depth (\cref{sec:setup}). ``--'' marks states a method does not estimate; best per-column values are bolded. The Tactile column repeats the Ours-8B prediction scores from \cref{tab:tactile_pred}; its predictions supply the contact mask used by Ours Full. $^{\dagger}$FP: FoundationPose~\cite{wen2024foundationpose}. Metric definitions: \cref{app:metrics}.}
\label{tab:component_benchmark_grouped}
\vspace{0.35em}
\centering
\footnotesize
\setlength{\tabcolsep}{3pt}
\resizebox{\textwidth}{!}{%
\begin{tabular}{@{}l cc cccc ccc cc@{}}
\toprule
\multirow{2}{*}{Method}
   & \multicolumn{2}{c}{Object}
   & \multicolumn{4}{c}{Hand}
   & \multicolumn{3}{c}{Body}
   & \multicolumn{2}{c}{Tactile} \\
\cmidrule(lr){2-3} \cmidrule(lr){4-7} \cmidrule(lr){8-10} \cmidrule(lr){11-12}
   & OTE$\downarrow$
   & ORE$\downarrow$
   & Rel-MPJPE$\downarrow$
   & PA-MPJPE$\downarrow$
   & GA-MPJPE$\downarrow$
   & Rel-ACC$\downarrow$
   & MPJPE$\downarrow$
   & PA-MPJPE$\downarrow$
   & Wrist$\downarrow$
   & Pearson$\uparrow$
   & CF1$\uparrow$ \\
\midrule
HaPTIC~\cite{haptic} + FP$^{\dagger}$
   & 286.8         & 120.7
   & 38.2          & 33.5          & 132.3         & 0.48
   & --            & --            & --
   & --            & --            \\
EgoEgo~\cite{li2023ego} + FP$^{\dagger}$
   & 286.8         & 120.7
   & --            & --            & --            & --
   & 232.0         & 144.5         & 609.8
   & --            & --            \\
\midrule
Ours w/o tactile
   & 286.8         & 120.7
   & \textbf{21.5} & \textbf{19.7} & \textbf{62.0} & \textbf{0.34}
   & \textbf{115.2}& \textbf{92.9} & \textbf{202.3}
   & --            & --            \\
\rowcolor{gray!15}
Ours Full
   & \textbf{171.4}& \textbf{57.8}
   & \textbf{21.5} & \textbf{19.7} & \textbf{62.0} & \textbf{0.34}
   & \textbf{115.2}& \textbf{92.9} & \textbf{202.3}
   & \textbf{0.812}& \textbf{0.816}\\
\bottomrule
\end{tabular}}
\vspace{-10pt}
\end{table*}

Following \cref{sec:setup,sec:recovery_pipeline}, \cref{tab:component_benchmark_grouped} reports egocentric HOI recovery on \ours{} under the fixed GT inputs. The first two rows are component-decoupled pipelines built from publicly available perception components: HaPTIC + FoundationPose covers hand and object but leaves body unmodeled, while EgoEgo + FoundationPose covers body and object but leaves hand unmodeled. Our pipeline (\cref{sec:recovery_pipeline}) supplies all three states simultaneously, with hand GA-MPJPE at $62.0$\,mm, body MPJPE at $115.2$\,mm, and wrist error at $202.3$\,mm.

The object metrics are identical for the first three rows because the same FoundationPose-based object stack is shared across all rows; predicted-tactile object-trajectory correction distinguishes Ours Full. With predicted tactile signals, OTE drops by $115.4$\,mm ($-40\%$) and ORE by $62.9^\circ$ ($-52\%$). The reported hand and body errors remain unchanged at the displayed precision.

\subsection{Humanoid Policy Learning}
\label{sec:exp_humanoid}

\begin{figure}[t]
  \centering
  \includegraphics[width=\linewidth]{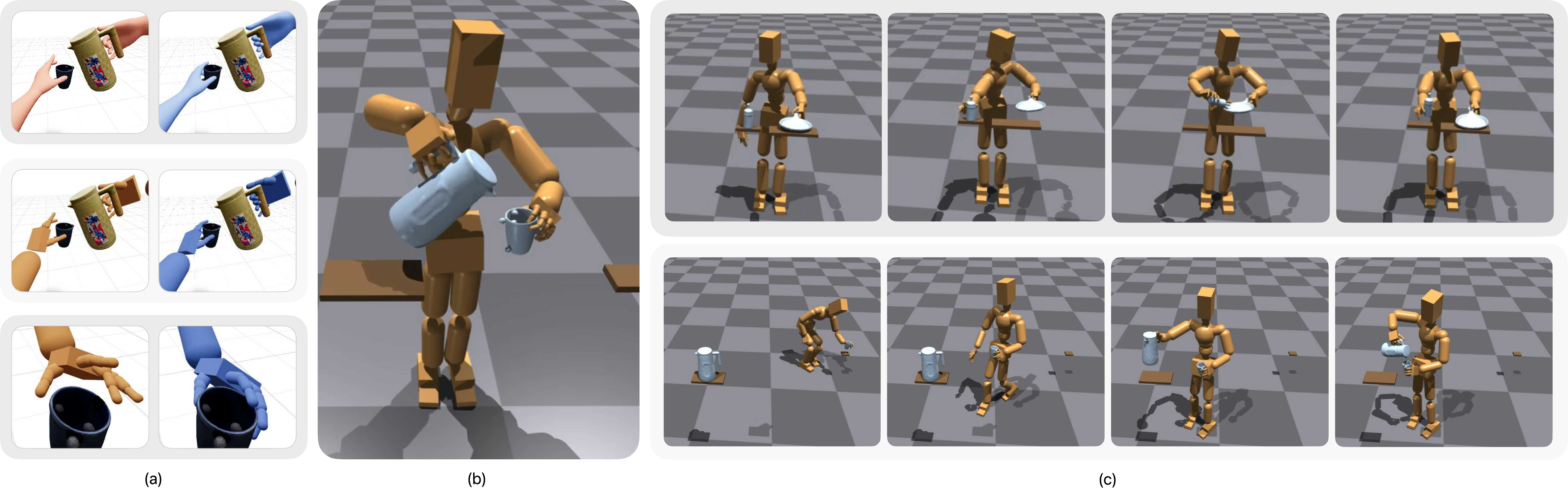}
  \caption{\textbf{Qualitative Humanoid Policy Rollouts.} (a) Raw (left) vs.\ tactile-refined (right) hand--object contacts and humanoid retargeting. (b) Whole-body bimanual pouring. (c) Policy rollouts on same-level (top) and cross-level loco-manipulation (bottom) tasks.}
  \label{fig:rl_policy}
\end{figure}

We further train whole-body dexterous manipulation policies for a simulated SMPL-X-based humanoid following InterMimic~\cite{xu2025intermimic}, using refined \ours{} demonstrations as supervision. During training, each policy is optimized using physics-based rewards that encourage both task completion and realistic hand-object contacts. In a paired comparison across $50$ reference motions with matched object trajectories, simulator, reward, and PPO configurations, policy success improves from $14/50$ ($28.0\%$) with raw MoCap demonstrations to $31/50$ ($62.0\%$) with demonstrations refined using measured tactile signals. The policies consistently reproduce key interaction patterns observed in the demonstrations, such as grasping, lifting, and manipulating objects at varying heights (\cref{fig:rl_policy}). These results demonstrate that \ours{} can effectively support downstream humanoid policy learning. Task examples and further discussion are in \cref{app:policy_learning}.

\FloatBarrier
\section{Conclusion}
\label{sec:conclusion}

In this work, we introduce \textbf{\ours{}}, a multimodal egocentric HOI benchmark designed to advance humanoid whole-body loco-manipulation from purely kinematic tracking toward physically grounded interaction. \ours{} provides synchronized whole-body motion, object trajectories, real tactile sensing, and egocentric RGB-D across diverse spatial layouts, enabling contact-aware evaluation and learning. Building upon this benchmark, we propose a tactile-guided refinement pipeline that leverages measured pressure as a grounding constraint to correct hand-object interactions while preserving global motion fidelity. This pipeline explicitly improves physical consistency, reducing the penetration-frame rate from $83.5\%$ to $4.2\%$. Furthermore, we establish the Egocentric HOI Recovery Benchmark, together with an egocentric tactile prediction model, demonstrating that contact evidence can be recovered from first-person observations and used to guide object-trajectory correction at recovery time. Overall, \ours{} provides a scalable and physically grounded foundation for studying whole-body dexterous interaction from egocentric human demonstrations. More importantly, it opens a practical path toward low-cost data collection for embodied intelligence: future systems may use lightweight egocentric sensors and predicted tactile evidence to recover contact-consistent demonstrations at scale, reducing reliance on expensive robot teleoperation or fully instrumented capture setups. We believe this benchmark can serve as a long-term testbed for connecting egocentric perception, tactile reasoning, and humanoid policy learning.

\textbf{Limitations \& Future Work.} While \ours{} provides rich demonstrations for whole-body loco-manipulation, our current humanoid policy evaluation remains in simulation, where discrepancies in sensor noise, actuator dynamics, and physical compliance between simulation and real hardware can affect direct deployment; future work will explore on-robot fine-tuning with real tactile feedback for closed-loop adaptation. For a controlled evaluation of HOI recovery, our benchmark currently uses ground-truth camera poses, object meshes, and depth, and focuses on rigid-body objects within our capture setting; estimating these geometric inputs, handling non-rigid objects, and learning policies directly from egocentrically recovered demonstrations remain future work. In addition, visual tactile prediction infers contact from visual cues rather than directly sensing latent physical properties such as mass or friction, making the integration of vision, proprioception, and real-time tactile sensing a key direction for real-world humanoid manipulation.

\begin{ack}
This work was supported by the State Key Laboratory of General Artificial Intelligence, the National Natural Science Foundation of China (62406195), the HPC Platform of ShanghaiTech University, the Core Facility Platform of Computer Science and Communication of ShanghaiTech University, the Key Laboratory of Intelligent Perception and Human-Machine Collaboration (ShanghaiTech University), Ministry of Education, and the Shanghai Engineering Research Center of Intelligent Vision and Imaging.
\end{ack}

\bibliographystyle{plainnat}
\bibliography{main}


\newpage
\appendix
\hypersetup{hidelinks}

\section{\ours{} Dataset Details}
\label{app:dataset_details}

This appendix expands the dataset construction summarized in \cref{sec:dataset}. We
describe the capture hardware, tactile representation, participants, and spatial
protocol. The provided annotations include synchronized SMPL-X body and hand motion,
rigid-object 6-DoF trajectories, camera poses, egocentric RGB-D, and per-frame
measured tactile signals.

\paragraph{Capture hardware.}
\ours{} is captured with $22$ OptiTrack Prime X 22 motion-capture cameras, a pair of Manus MetaGloves for fine hand motion, a pair of custom tactile gloves, each with $256$ sensing points, and an Orbbec EGO Quad four-camera headband. Egocentric depth is computed from the stereo RGB views. The exported modalities are temporally aligned at $30$\,Hz, with human and object motion expressed in a shared coordinate system.

\paragraph{Tactile signals and SMPL-X hand keypoints.}
\label{app:tactile_representation}
The tactile measurements are mapped to a $32$-dimensional representation associated
with selected SMPL-X hand keypoints. Measured tactile signals
guide human-motion refinement during dataset construction (\cref{sec:refine}) and
provide supervision for tactile prediction (\cref{sec:tactile_pred}).

\paragraph{Participants and spatial protocol.}
\ours{} contains data from six participants, four male and two female,
with varying heights and body proportions. The five interaction levels---Floor, Knee,
Waist, Chest, and Neck---are calibrated to each participant's height rather than fixed
at the same absolute heights for all participants. Together with three horizontal
variations, the body-calibrated grid exposes changes in posture, reach, and viewpoint
across interaction types.

\paragraph{Collection stages.}
The three collection stages of \cref{sec:overview} are deployed as follows. \textbf{S1
(same-level interactions)} records each object--action pairing within a single layout
cell and is the primary source of controlled variation across the body-calibrated
grid. \textbf{S2 (cross-level transitions)} records HOI motion that includes
locomotion, by carrying objects across multiple height levels. \textbf{S3
(long-horizon scene-level activity)} chains multiple interactions and transitions into
continuous activities across freely arranged scene layouts. S3 additionally provides
synchronized exocentric views recorded with four Z~CAM cameras for independent
qualitative inspection of body motion, object trajectories, and interaction transitions;
these views are not recovery inputs.

\begin{figure}[htbp]
  \centering
  \includegraphics[width=\linewidth]{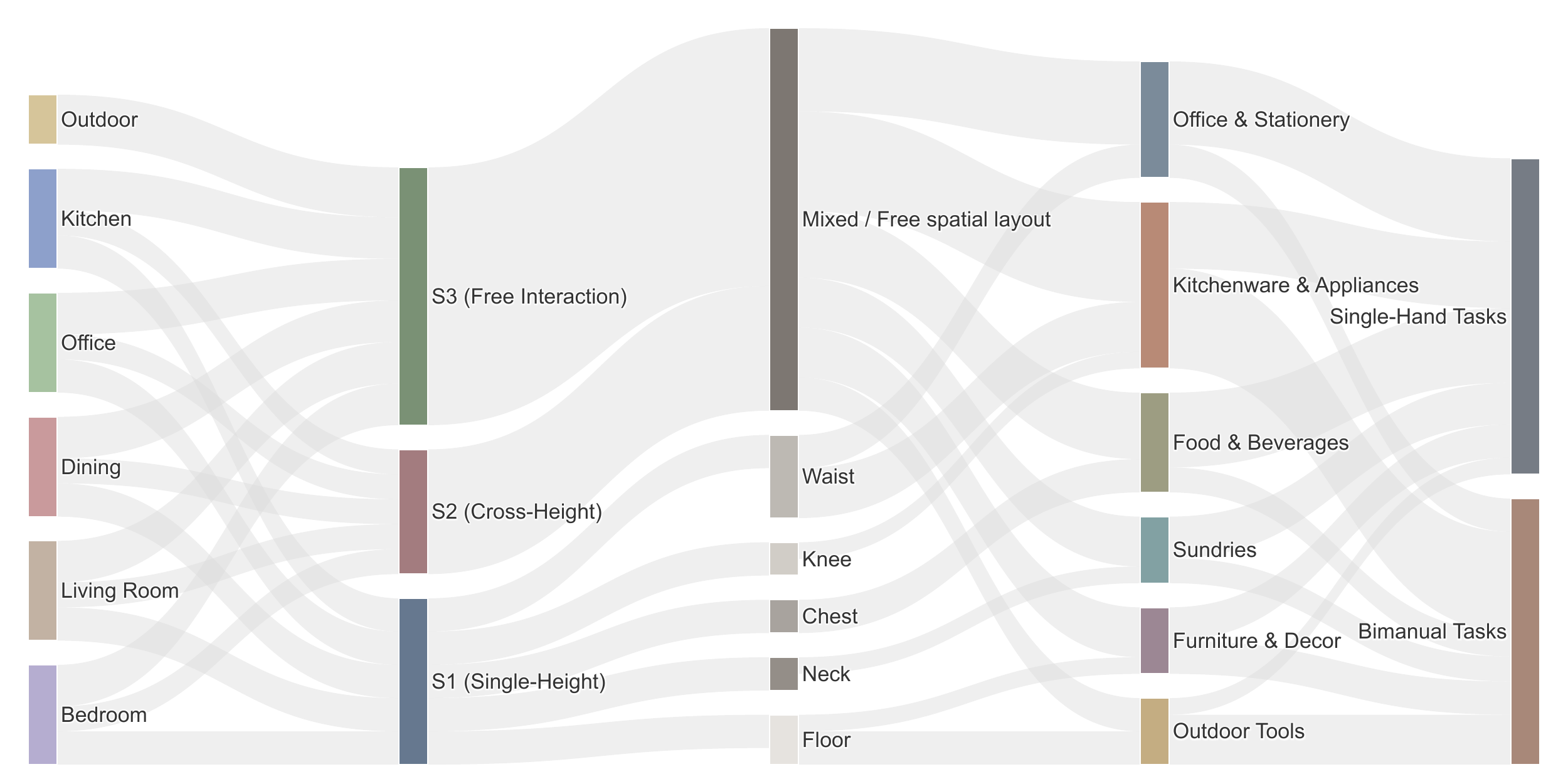}
  \caption{Scenes, collection stages, spatial layouts, object categories, and hand modes in \ours{}.}
  \label{fig:hoi_sankey}
\end{figure}

\paragraph{Interaction coverage.}
\ours{} covers six everyday scenes and $52$ objects, with both single-handed and bimanual activities. It includes prehensile, in-hand, and non-prehensile behaviors, such as grasping and transport, object reorientation during use, pushing, pouring, and bimanual cooking activities, embedded within long-horizon whole-body interactions. The object annotations describe rigid-body motion (\cref{fig:hoi_sankey}).

\FloatBarrier

\begin{figure}[htbp]
  \centering
  \IfFileExists{Figures/demo1.pdf}{%
    \includegraphics[
      width=\linewidth,
      height=0.72\textheight,
      keepaspectratio
    ]{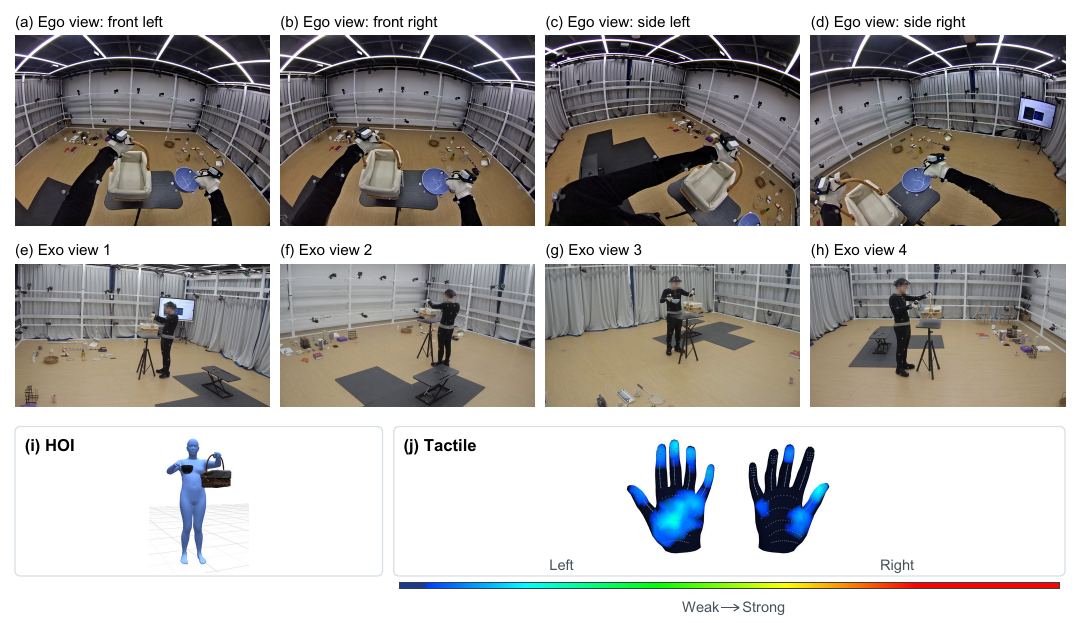}%
  }{%
    \fbox{\parbox[c][0.28\textheight][c]
      {\dimexpr\linewidth-2\fboxsep-2\fboxrule\relax}{%
        \centering\texttt{Figures/demo1.pdf}
    }}%
  }
  \caption{Synchronized egocentric and exocentric views, SMPL-X HOI,
  and tactile visualizations in \ours{}.}
  \label{fig:multimodal_demo1}
\end{figure}

\begin{figure}[htbp]
  \centering
  \IfFileExists{Figures/demo2.pdf}{%
    \includegraphics[
      width=\linewidth,
      height=0.72\textheight,
      keepaspectratio
    ]{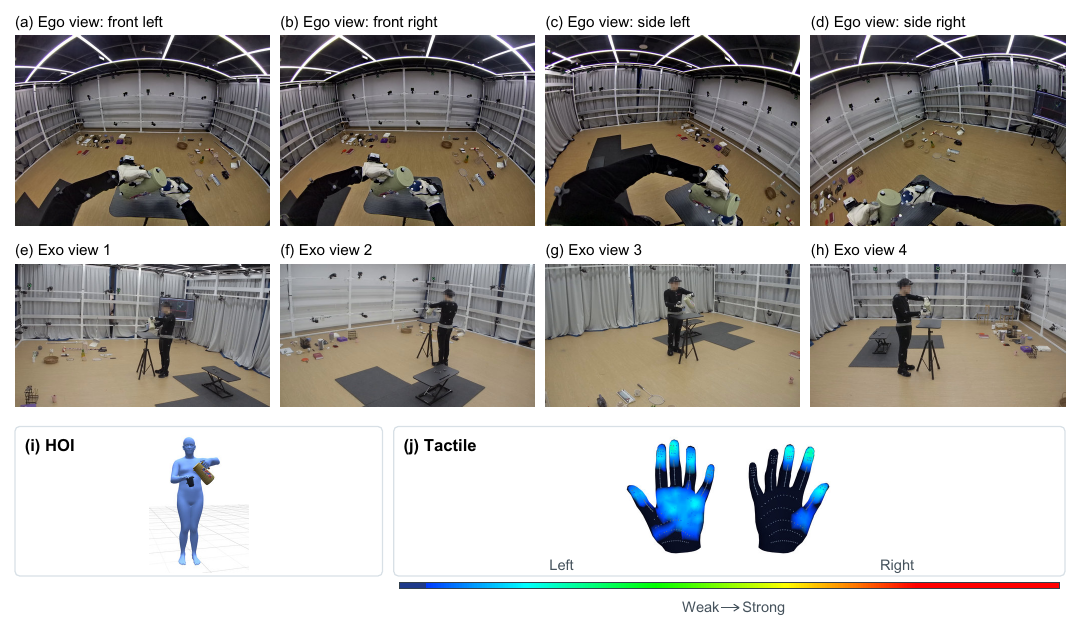}%
  }{%
    \fbox{\parbox[c][0.28\textheight][c]
      {\dimexpr\linewidth-2\fboxsep-2\fboxrule\relax}{%
        \centering\texttt{Figures/demo2.pdf}
    }}%
  }
  \caption{An additional synchronized multimodal example from \ours{}.}
  \label{fig:multimodal_demo2}
\end{figure}

\paragraph{Synchronized examples.}
\Cref{fig:multimodal_demo1,fig:multimodal_demo2} show synchronized
egocentric and exocentric views, SMPL-X HOI, and tactile visualizations,
with the complete object collection displayed in the background.

\FloatBarrier

\section{Tactile Refinement and Prediction}
\label{app:tactile_details}

\subsection{Measured-Tactile Human-Motion Refinement}
\label{app:tactirefine_full}

Body and object trajectories from MoCap are reliable, while local hand-object contact
is the most fragile part of whole-body interaction data. We therefore anchor
human-motion refinement in measured tactile pressure. During dataset construction,
human motion is refined while the MoCap object trajectory remains fixed.

\paragraph{Interaction interval detection.}
The dual tactile-geometric gate of \cref{sec:refine} flags a frame when the
maximum aggregated finger pressure exceeds $\tau_p$ and the hand-mesh-to-object
distance falls below $\tau_d$. Short isolated intervals are
removed, and nearby intervals are merged before template selection. A representative
template frame is selected using tactile confidence, kinematic stability, and
geometric proximity, defining the wrist-relative reference for the subsequent steps.

\paragraph{Pressure-modulated local refinement.}
At the template frame, hand articulation and wrist pose are optimized under an
objective whose weights derive from measured fingertip tactile signals. For each
finger, pressure readings are aggregated over a temporal neighborhood and normalized
to form a continuous tactile weight. These weights modulate attraction between sparse
hand contact points and object-surface targets, while a soft pose-preservation term
penalizes deviations from the captured hand articulation. The attraction is introduced
in stages.

\paragraph{Spatiotemporal propagation and collision refinement.}
The refined wrist-relative transform is composed with the object trajectory to obtain
a target wrist trajectory. The wrist target and hand-pose template are blended with
the captured motion through smooth weights that ramp inside each interaction interval
and decay at its boundaries. Body posture is adjusted through bounded inverse
kinematics to follow the refined wrist target, with the root pose fixed. With the
wrist trajectory fixed, the collision-aware step refines finger pose on near-contact
frames. The penetration term penalizes hand vertices inside the object, while the
closeness term anchors the result to the blended pose (\cref{eq:collision}).
The wrist-relative propagation assumes approximately stable contact within each
interval.

\subsection{Tactile Prediction from Egocentric Video}
\label{app:egotacti}

\paragraph{Input and backbone.}
The predictor consumes a short egocentric video clip together with a fixed
task-conditioning prompt. We use
Qwen3-VL~\citep{Qwen3-VL} 2B and 8B as the multimodal backbone. LoRA adapters are
injected into the language-layer attention projections, with rank $r{=}16$
and scaling $\alpha{=}32$. The adapters, visual-language
merger, and regression head are trained, while the remaining backbone weights are
frozen.

\paragraph{Regression output.}
Inspired by OpenVLA~\citep{kim24openvla}, we attach a small two-layer MLP regression
head to the backbone. The head is supervised by synchronized measured tactile signals
and predicts non-negative values for the target-frame representation defined in \cref{app:tactile_representation}.
The predictor learns visual interaction cues for human-demonstration collection;
it does not explicitly estimate physical properties such as mass, friction, or
fill level. Robots with reliable tactile sensors can use measured feedback during
execution.

\paragraph{Regression and contact supervision.}
Weighted mean-squared error supervises the tactile-signal magnitude.
Non-zero targets are up-weighted to counter contact sparsity. Binary
cross-entropy supervises contact masks derived from a fixed pressure
threshold.

\paragraph{Supervised baseline.}
The DINOv2~\citep{oquab2023dinov2} ViT-B/14 baseline uses a frozen visual encoder with a learned temporal
prediction head. The baseline and Ours use the same training data, egocentric RGB
inputs, measured tactile supervision, and evaluation metrics. This comparison
complements the zero-shot baselines with a learned predictor of the tactile signal.

\paragraph{Zero-shot baselines.}
The zero-shot Qwen3-VL baselines use native autoregressive text generation with fully
frozen models, without LoRA adapters or a learned regression head. A structured prompt
produces ordinal contact scores, which are calibrated and mapped to the same tactile
representation for evaluation. All samples are retained: missing or invalid fields are
set to zero, and outputs with no valid fields are scored as all-zero predictions. The
zero-shot baselines and Ours are evaluated against the same measured signals using
the same metrics.

\paragraph{Use of predicted tactile signals.}
At recovery time, predicted tactile signals identify contact intervals for
post-processing. The recovered body and wrist trajectories are held fixed, while
finger poses and the object trajectory are adjusted. The measured-pressure
weighting described in \cref{app:tactirefine_full} is used during dataset
construction and is not applied to predicted tactile values.

\section{Evaluation Metric Definitions}
\label{app:metrics}

Recovery is evaluated using GT camera poses and GT object meshes, with wrist
positions lifted using GT depth. Hand and body errors
are measured against the measured-tactile-refined MoCap motion, while object errors
are measured against the MoCap object trajectories. The following definitions
distinguish recovery accuracy from changes introduced during dataset refinement.

\paragraph{Hand metrics.}
\textbf{Rel-MPJPE} subtracts the wrist root per frame before averaging the per-joint Euclidean error. \textbf{PA-MPJPE} applies per-frame Procrustes alignment between predicted and GT joints before computing MPJPE. \textbf{GA-MPJPE} aligns each evaluated trajectory once via a global rigid transform and reports MPJPE on the aligned joints. \textbf{Rel-ACC} reports the mean root-relative joint-acceleration error.

\paragraph{Object metrics.}
\textbf{OTE} is the per-frame translation error in millimeters between predicted and GT object centroids. \textbf{ORE} is the per-frame angular error in degrees between predicted and GT object orientations, measured as the geodesic distance on $SO(3)$. Both are averaged over the evaluation frames of each object.

\paragraph{Body metrics.}
\textbf{MPJPE} is the per-frame Euclidean joint error in millimeters on the common body-joint subset. \textbf{PA-MPJPE} applies per-frame Procrustes alignment before MPJPE. \textbf{Wrist} is the world-frame body-FK position error averaged over the two wrist joints.

\paragraph{Refinement-ablation metrics.}
For the refinement ablation in \cref{tab:postopt}, \textbf{PR} is the fraction of
frames containing any penetrating hand vertex. \textbf{FP-C} is computed over
frame--finger pairs as the fraction where the hand is in geometric near-contact with
the object but the corresponding measured tactile signal is inactive. Near-contact
denotes penetration or an unsigned surface distance below $30$\,mm. \textbf{NTF-D} is
the mean displacement in millimeters of fingers without tactile evidence relative to
the raw capture. \textbf{HPD} is the mean per-joint hand-pose deviation in radians
relative to the raw capture. The refinement-ablation \textbf{MPJPE} reports full-body
joint displacement against the captured motion. These displacement measures quantify
changes to the captured demonstration, whereas PR and FP-C describe geometric and
contact consistency.

\paragraph{Tactile prediction metrics.}
\textbf{Pearson} measures correlation between predicted and measured tactile signals. \textbf{MAE} and \textbf{MSE} report absolute and squared prediction errors in the processed tactile-signal space. \textbf{CF1} is element-level contact F1, computed from predicted and measured contact masks using a threshold of $0.5$. \textbf{FC@$k$} is frame-level contact F1: a frame is labeled as contact when any of its tactile channels exceeds pressure threshold $k$ on the processed tactile-signal scale. \textbf{R$^2$} is computed after flattening the evaluated frames and tactile channels within each sequence, and the resulting scores are averaged equally across sequences. Together, these metrics assess tactile-signal agreement and contact detection.

\section{Additional Experimental Results}
\label{app:additional_results}

\paragraph{Supervised comparison.}
In the main tactile-prediction evaluation, Ours-8B outperforms the
supervised DINOv2 ViT-B/14 baseline, improving Pearson correlation from
$0.674$ to $0.812$ and reducing MAE from $1.994$ to $1.449$ and MSE
from $12.45$ to $7.12$.

\paragraph{Generalization to unseen subjects and objects.}
\label{app:generalization}
We further evaluate tactile prediction on jointly unseen subjects and objects. Both
models are evaluated without retraining or adaptation. Ours-8B achieves a Pearson
correlation of $0.795$, compared with $0.812$ in the main evaluation, and
outperforms the supervised DINOv2 ViT-B/14 baseline on all reported metrics
(\cref{tab:joint_holdout}). These results concern unseen subjects and objects within
our capture setting; controlled physical-property shifts and in-the-wild evaluation
remain future work.

The unseen objects introduce variation in appearance, geometry, material, surface
characteristics, and manipulation affordances. Examples include stone
mortar-and-pestle tools, a transparent glass kettle, fabric-covered furniture, a metal
cup, and wooden building blocks. These objects also involve distinct grasp/contact
configurations and manipulation patterns, such as grinding and block assembly, rather
than only new instances with minor visual changes.

\begin{table}[htbp]
  \centering
  \small
  \setlength{\tabcolsep}{5pt}
  \caption{Tactile prediction on jointly unseen subjects and objects. Metric definitions are given in \cref{app:metrics}.}
  \label{tab:joint_holdout}
  \begin{tabular*}{\linewidth}{@{\extracolsep{\fill}}lccccccc@{}}
    \toprule
    Method & Pearson$\uparrow$ & CF1$\uparrow$ & FC@3$\uparrow$ & FC@5$\uparrow$ & MAE$\downarrow$ & MSE$\downarrow$ & R$^2$$\uparrow$ \\
    \midrule
    DINOv2 ViT-B/14 & 0.485 & 0.645 & 0.740 & 0.715 & 3.581 & 23.90 & 0.22 \\
    Ours-8B & \textbf{0.795} & \textbf{0.757} & \textbf{0.816} & \textbf{0.936} & \textbf{1.891} & \textbf{9.42} & \textbf{0.61} \\
    \bottomrule
  \end{tabular*}
\end{table}

\paragraph{Whole-body dexterous policy learning.}
\label{app:policy_learning}
\ours{} targets interactions that require coordinated whole-body control across different heights and precise bimanual manipulation. For example, in a low-height kettle-pouring motion, the humanoid must maintain a squat, insert one hand into a narrow kettle handle and control the pouring motion, while using the other hand to grasp and stabilize the cup. Small errors in body balance, wrist pose, or finger placement can break the required contacts and cause failure. We follow InterMimic~\citep{xu2025intermimic} for simulated SMPL-X-based humanoid policy learning and evaluation. In a paired comparison across $50$ reference motions (training $100$ policies for $2{,}000$ epochs each with matched object trajectories, simulator, reward, and PPO configurations), a motion is counted as successful if reproduced at least once across $10$ trials. Measured-tactile refinement improves policy success from $14/50$ ($28.0\%$) with raw MoCap demonstrations to $31/50$ ($62.0\%$) with refined demonstrations. This comparison assesses the learnability of measured-tactile-refined MoCap demonstrations; policy learning directly from egocentrically recovered demonstrations remains future work.

\end{document}